\documentclass[journal]{IEEEtran}
\newcommand{\fref}[1]{Figure \ref{#1}}
\newcommand{\sref}[1]{Section \ref{#1}}
\newcommand{\tref}[1]{Table \ref{#1}}
\newcommand{\eref}[1]{Eq. (\ref{#1})}

\newcommand{\alref}[1]{Algorithm \ref{#1}}
\usepackage{graphicx}
\usepackage{amsmath}
\usepackage{amsthm}
\newtheorem{definition}{Definition}
\usepackage{amssymb,amsfonts}
\usepackage{algorithmicx}
\usepackage{array}
\usepackage{multirow}
\usepackage{mathrsfs}%
\usepackage{xcolor}%
\usepackage{textcomp}%
\usepackage{manyfoot}%
\usepackage{booktabs}%
\usepackage{algorithm}%
\usepackage{algpseudocode}%
\usepackage{listings}%
\usepackage{hyperref}
\usepackage{balance}
\usepackage{tabularx}
\usepackage{enumitem}
\setlist[enumerate]{label={(\arabic*)}}
\usepackage{soul} 
\usepackage{array}
\usepackage{stfloats}
\usepackage{url}
\usepackage{verbatim}
\usepackage{epsfig}
\usepackage{color}
\usepackage{colortbl}
\usepackage{lineno}
\usepackage{graphics}
\usepackage{lmodern}
\graphicspath{{figures/}}
\usepackage{rotating}
\usepackage{float}
\usepackage{cases}
\usepackage{supertabular}
\usepackage{makecell}
\ifCLASSOPTIONcompsoc
 \usepackage[caption=true,font=normalsize,labelfont=sf,textfont=sf]{subfig}
\else
 \usepackage[caption=true,font=footnotesize]{subfig}
\fi
\usepackage{url}

\begin{document}
%
\title{Reinforcement learning Based Automated Design of Differential Evolution Algorithm for Black-box Optimization}
\author{Xu~Yang,
        Rui~Wang,
        Kaiwen~Li,
        and~Ling~Wang,
\thanks{Xu Yang and Rui Wang are with the Department
of Systems Engineering, National University of Defense Technology, Changsha 410073, China (e-mail: yangxunudt@outlook.com; rui\_wang@nudt.edu.cn; likaiwen@nudt.edu.cn).}
\thanks{Ling Wang is with the Department of Automation, Tsinghua University, Beijing 100084, China (e-mail: wangling@tsinghua.edu.cn).}
\thanks{Manuscript received April 19, 2024; revised September 17, 2024. (\textit{Corresponding author: Rui Wang})}}

\markboth{IEEE,~Vol.~X, No.~X, X~X}%
{Shell \MakeLowercase{\textit{et al.}}: Bare Demo of IEEEtran.cls
for Journals}

\maketitle

\begin{abstract}
  Differential evolution (DE) algorithm is recognized as one of the most effective evolutionary algorithms, demonstrating remarkable efficacy in black-box optimization due to its derivative-free nature. Numerous enhancements to the fundamental DE have been proposed, incorporating innovative mutation strategies and sophisticated parameter tuning techniques to improve performance. However, no single variant has proven universally superior across all problems. To address this challenge, we introduce a novel framework that employs reinforcement learning (RL) to automatically design DE for black-box optimization through meta-learning. RL acts as an advanced meta-optimizer, generating a customized DE configuration that includes an optimal initialization strategy, update rule, and hyperparameters tailored to a specific black-box optimization problem. This process is informed by a detailed analysis of the problem characteristics. In this proof-of-concept study, we utilize a double deep Q-network for implementation, considering a subset of 40 possible strategy combinations and parameter optimizations simultaneously. The framework's performance is evaluated against black-box optimization benchmarks and compared with state-of-the-art algorithms. The experimental results highlight the promising potential of our proposed framework.
\end{abstract}


\section{Introduction}


\IEEEPARstart{B}{lack-box} optimization (BBO) is becoming increasingly crucial for tackling complex real-world optimization challenges, where traditional methods often falter due to their reliance on detailed mathematical models. Differential Evolution (DE), a population-based evolutionary algorithm, has gained widespread recognition for its effectiveness in addressing black-box optimization problems, leading to the development of numerous variants \cite{tan2021differential}.

However, in accordance with the \textit{No Free Lunch} theorem \cite{wolpert1997no}, no single algorithm can consistently outperform all others across every problem instance. Additionally, the dynamic nature of many real-world problems demands algorithms capable of rapidly adapting to evolving optimization landscapes. Manually designing an algorithm tailored to a specific problem requires extensive expert knowledge, while selecting an efficient algorithm from thousands of candidates is highly time-consuming. To address these issues, meta-optimizers that can automatically select, configure, or even generate algorithms tailored to specific problems \cite{reddy2022self, chen2022adaptive} have emerged as a promising methodology \cite{WU2019695}. Meta-optimizers have driven evolutionary algorithms capable of customizing their own evolutionary strategies for a given problem automatically.

One of the most widespread implementations of meta-optimizers focuses on machine learning techniques, particularly reinforcement learning (RL) \cite{SONG2024101517}, which has facilitated the development of algorithms that can learn from given problems, resulting in more efficient and effective DEs. Studies \cite{tan2021differential, li2019differential, fister2022reinforcement, li2023scheduling} have focused on RL-assisted self-adaptive operator selection of DE. Li et al. \cite{li2019differential} treated each individual in the population as an agent, using fitness rankings to encode hierarchical state variables and employing three typical DE mutation strategies as optional actions for the agent. Tan et al. \cite{tan2021differential} implemented adaptive selection of mutation strategies in the DE evolution process based on a deep Q-network (DQN). Fister et al. \cite{fister2022reinforcement} utilized reinforcement learning to select strategies derived from the original L-SHADE \cite{tanabe2014improving} algorithm, transitioning from the 'DE/current-to-pbest/1/bin' mutation strategy to the iL-SHADE \cite{brest2016shade} and jSO \cite{brest2017single} using the 'DE/current-to-pbest-w/1/bin' mutation strategies. Li et al. \cite{li2023scheduling} proposed an adaptive multi-objective differential evolutionary algorithm based on deep reinforcement learning to effectively obtain Pareto solutions, integrating RL as a controller that can adaptively select mutation operators and parameters according to different search domains.

Several attempts have also been made to employ RL to configure EAs. Huynh et al. \cite{huynh2021q} and Peng et al. \cite{peng2023reinforcement} used Q-learning to control predetermined parameters. Sun et al. \cite{sun2021learning}, Zhang et al. \cite{zhang2022variational}, Liu et al. \cite{liu2023learning}, and Zhang et al. \cite{zhang2023reinforcement} employed policy-based RL for more precise parameter control. It can be concluded that these aforementioned studies harnessed the distribution characteristics of the population. In addition, it is noteworthy that the majority of these investigations focused solely on mutation strategies or numerical parameters.

To address these limitations, we introduce an innovative framework that employs RL as a meta-optimizer to generate DE tailored to the characteristics of specific problems. In this framework, meta-learning \cite{hospedales2021meta} is incorporated to enhance the generalizability of the meta-optimizer, thereby enabling it to adapt more effectively across a diverse range of problem scenarios. Specifically, problem characteristics are identified via exploratory landscape analysis before evolution and encoded as the state. The action can be decoded from selections regarding different components and parameters to create a tailored DE. The RL agent is trained offline by meta-learning pairs $(state, action, reward) \rightarrow (characteristics, DE, performance)$ across a large number of black-box optimization problems (BBOPs). Subsequently, the agent can autonomously generate a bespoke DE tailored for unseen problems. For the implementation of this framework, we utilize a double deep Q-network (DDQN) \cite{mnih2015human}.

Contrary to other mainstream methods, such as hyper-heuristic methods \cite{burke2013hyper}, which conduct searches prior to the optimization process to produce an adequate combination of provided components for a given specific problem, and self-adaptive methods \cite{kramer2010evolutionary, frank2013selfadaptive}, which perform algorithm configuration during evolution, our proposed method directly generates bespoke DE with different components for targeted problems without additional search and before evolution.

The reminder of this work is structured as follows. Preliminary works including DE versus RL are introduced in \sref{preliminary}. Exploratory landscape analysis used in rlDE is provided in \sref{ela}. Detailed rlDE is shown in \sref{rleag}. Experimental setup, results and analysis are provided in \sref{exp}. \sref{conclusion} summarizes this work and discusses future work.

\section{Preliminary}
\label{preliminary}
DE and RL are two powerful methodologies in the domains of optimization and machine learning, respectively. Each possesses unique strengths and weaknesses, and their combination through hybrid algorithms can yield more robust and efficient solutions. Given the integration of both approaches in our proposed method, this section provides a brief overview of the fundamentals of DE and RL, as well as a review of hybrid methods currently suggested in research, to facilitate a clearer understanding.

\subsection{DE}
DE was introduced by Storn and Price \cite{storn1997differential}. As one of the most popular evolutionary algorithms (EAs), DE is a population-based optimization technique inspired by the process of natural evolution \cite{back1996evolutionary}. It iteratively evolves a population of candidate solutions by generating offspring and selecting the best-performing individuals. Following the scheme of EAs, DE variants mainly focus on various mutation strategies, such as "DE/rand/1", "DE/rand/2", "DE/best/1", "DE/best/2", "DE/current-to-best/1", and "DE/current-to-rand/1" \cite{fan2015self}. The general process of DE is illustrated in Algorithm \ref{algDE}. In DE literature, a parent vector in the population $P$ from the current generation is referred to as the target vector, a mutant vector obtained through the differential mutation operation is known as the donor vector, and an offspring formed by recombining the donor with the target vector is called the trial vector.

\begin{algorithm} 
  \caption{Differential evolution algorithm} 
  \begin{algorithmic}[1] 
    \State $P$ $\leftarrow$ apply initialization strategy 
    \While{termination condition not met} 
    \State $target$ $\leftarrow$ $P$
    \State $donor$ $\leftarrow$ apply mutation strategy on $target$ 
    \State $trial$ $\leftarrow$ apply crossover strategy on $(donor,target)$ 
    \State $P$ $\leftarrow$ apply selection strategy on $(target, trial)$
    \EndWhile 
  \end{algorithmic} 
  \label{algDE}
\end{algorithm}

\subsection{RL}
RL is a type of machine learning designed for sequential decision-making problems, where an agent learns to make decisions by taking actions in an environment to maximize cumulative reward \cite{sutton2018reinforcement}. RL is formalized as a Markov Decision Process (MDP), denoted as a tuple $(\mathcal{S}, \mathcal{A}, P, R, \gamma)$, where:
\begin{definition}{MDP $(\mathcal{S}, \mathcal{A}, P, R, \gamma)$}
\begin{itemize}
    \item $\mathcal{S}$ is the set of states,
    \item $\mathcal{A}$ is the set of actions,
    \item $P$ is the transition probability function,
    \item $R$ is the reward function assigning a numerical reward to each state-action pair,
    \item $\gamma$ is the discount factor.
\end{itemize}
\end{definition}

The objective of training an RL agent is to learn a policy $\pi$ that maps states to actions, in order to maximize the expected cumulative reward:
\begin{equation}
  J(\pi) = \mathbb{E}\left[\sum_{k=0}^{\infty} \gamma^k R_{t+k+1} \mid \pi, s_t\right]
\end{equation}
where $s_t$ is the state at time $t$, and $R_{t+k+1}$ is the reward at time $t+k+1$.

RL can be categorized into different families based on their approach to learning:
\begin{itemize}
  \item \textbf{value-based RL}: Learn value functions to estimate the expected future reward and derive optimal policies from them. Examples include Q-Learning \cite{clifton2020q}, DQN \cite{mnih2015human} and state-action-reward-state-action (SARSA) \cite{rummery1994line}. 
  \item \textbf{policy-based RL}: Directly optimize the policy function to maximize the expected reward. Examples include Policy Gradient \cite{sutton1999policy} and Actor-Critic methods \cite{grondman2012survey}.
  \item \textbf{model-based RL}: Learn a model of the environment's transition dynamics to simulate future states and actions, enabling efficient planning and exploration. Examples include Model Predictive Control \cite{arroyo2022reinforced} and Monte Carlo Tree Search \cite{mo2021safe}.
\end{itemize}

\subsection{RL-assisted DE}
To overcome the limitations of DE, researchers have proposed hybrid methods that integrate the decision-making capabilities of RL into DE. The RL agent is utilized to guide the search process of DE \cite{tan2021differential, song2024reinforcement, li2023scheduling, huynh2021q}. Li and Hao et al. have conducted an in-depth analysis of each research direction, organizing multiple research branches \cite{li2024bridging}. \tref{erl} lists the main contributions on DE in the past decade.

\begin{table*}
  \centering
  \caption{Main contributions of RL-assisted DE in the past decade} 
  \begin{tabular}{cccc}
    \toprule
    {\textbf{Category}} & \textbf{References} & \textbf{RL} & \textbf{Optimization object}\\
    \hline
    \multirow{4}{*}{operator selection} & DEDQN \cite{tan2021differential} & DQN &six mutation strategies\\
     &DEDDQN \cite{sharma2019deep} & DDQN & mutation strategy\\  
     &RL-HDE \cite{peng2023reinforcement} &Q-learning &mutation strategy, trigger parameters\\
     & DE-RLFR \cite{li2019differential}  &Q-learning &mutation strategy\\
     \hline
     \multirow{4}{*}{hyperparameter configuration} & Q-LSHADE \cite{zhang2023controlling} & Q-learning & trigger parameters\\
     & REM \cite{zhang2022variational} & VPG & scale factor, crossover rate\\
     & qlDE \cite{huynh2021q} & Q-learning & scale factor, crossover rate \\
     & RLDE \cite{hu2021reinforcement} & Q-learning  & crossover rate\\

    \bottomrule
  \end{tabular}
  \label{erl}
\end{table*}

While these methods have been demonstrated effective on various test suites, they still exhibit common limitations. On one hand, these studies often employ RL as an adaptive operator during evolution, which may increase computational cost. On the other hand, the optimization objects in each study of \tref{erl} typically focus on only one type, either discrete or continuous.

\section{Exploratory landscape analysis}
\label{ela}
A comprehensive understanding of the BBOPs' characteristics offers valuable insights into their complexity and guides the selection of the most appropriate optimization algorithm \cite{vskvorc2020understanding}. While certain high-level characteristics, such as multi-modality and global structure, are used to describe BBOPs, identifying these features without expert knowledge can be challenging. To address this issue, ELA provides a structured methodology for characterizing low-level features \cite{hanster2017flaccogui}, which has been shown to be effective in classifying BBOPs \cite{renau2021towards} and informative for algorithm selection \cite{kerschke2019automated}. This section introduces a discussion of the used features in our framework that are computationally inexpensive to calculate.

In recent years, researchers have been expanding the set of low-level features (\cite{prager2024exploratory,derbel2019new,kerschke2014cell,munoz2014exploratory}), which are distributed across various platforms and use different feature collections. Without loss of generality, it is necessary to use multiple feature collections at once for a careful characterization while single feature collections measure few properties. Pflacco package \cite{prager2023pflacco} addresses this issue, which contains more than 300 features distributed across 17 feature sets. 

\begin{enumerate}
  \item Initial ELA Features (Convexity, Curvatyre, $y$-distribution, Meta-model, Local Search, Levelset)
  \item Cell Mapping (Angle, Convexity and Gradient Homogeneity) and Generakized Cell Mapping Features
  \item Barrier Tree Features
  \item Information Content Features 
  \item Disperison Features
  \item Miscellaneous Approaches (Basic, Linear Models, Principal Components)
\end{enumerate}

Each group consists of a set of sub-features that can be computed based on the initial sample data $D^s = \{X,y\}$, where $X$ represents the decision vector with dimension $D$, and $y$ represents the observation value vector. 

Among these feature sets, some cheap features from six sets are chosen to characterize BBOPs in our framework following the discussion in \cite{renau2019expressiveness}. 

\begin{equation}
  \text{Skewness} = \frac{\sqrt{n} \sum_{i=1}^{n} (y_i - \bar{y})^3}{\left( \sum_{i=1}^{n} (y_i - \bar{y})^2 \right)^{3/2}} \cdot \left(1 - \frac{1}{n}\right)^{3/2}
  \label{ske}
\end{equation}

\begin{equation}
  \text{Kurtosis} = \left( \frac{\sum_{i=1}^{n} (y_i - \bar{y})^4}{n} \right) \left( \frac{1}{\sum_{i=1}^{n} (y_i - \bar{y})^2} \right)^2 - 3
  \label{kur}
\end{equation}

For $y$-distribution, skewness, kurtosis, and the number of peaks are computed. Skewness measures the asymmetry of the probability distribution of a real-valued random variable about its mean, as given by \eref{ske}. Kurtosis quantifies the "tailedness" of the probability distribution of a real-valued random variable, as shown in \eref{kur}. The number of peaks is determined by identifying the local maxima in the distribution of the data. The Gaussian Kernel Density Estimation (KDE) is used. The general steps are:

\begin{itemize}
  \item Use KDE to estimate the probability density function of the data, denoted as $\hat{f}(x)$.
  \item Define the bounds of the interval, which in this study is set as $interval = [min(y)-3*\lambda*std(y), max(y)+3*\lambda*std(y)]$, where $\lambda$ is a scaling factor used to determine the kernel's bandwidth. 
  \item Evaluate $\hat{f}(x)$ over a grid of points within $interval$.
  \item Identify local minima of $\hat{f}(x)$ and partition the interval into regions between consecutive minima.
  \item For each region, calculate the mean of $\hat{f}(x)$ plus the absolute difference between the positions of the region's endpoints.
  \item Count the number of regions where this value exceeds 0.1, denoted as $n_{peaks}$.
\end{itemize}

For Levelset, classification methods such as Linear Discriminant Analysis (LDA) and Quadratic Discriminant Analysis (QDA) are considered. The related feature values include the Mean Misclassification Error (MMCE) of LDA and QDA (denoted as $MMCE_{LDA, q}$ and $MMCE_{QDA,q}$), and the ratio $r_q$ of $MMCE_{LDA, q}$ and $MMCE_{QDA,q}$, where $q$ represents the quantiles which is set to $[0.1, 0.25, 0.5]$. The calculation is shown in \eref{ldaq}, \eref{qdaq} and \eref{rq}. 

\begin{equation}
  MMCE_{LDA, q} = \frac{1}{n} \sum_{i=1}^{n} (y_{class, test_i} \neq \hat{y}_{LDA, test_i})
  \label{ldaq}
\end{equation}

\begin{equation}
  MMCE_{QDA, q} = \frac{1}{n} \sum_{i=1}^{n} (y_{class, test_i} \neq \hat{y}_{QDA, test_i})
  \label{qdaq}
\end{equation}

\begin{equation}
  r_q = \frac{MMCE_{LDA, q}}{MMCE_{QDA, q}}
  \label{rq}
\end{equation}
where $y_{class, test_i}$ is the true classification of the test set while $\hat{y}_{*, test_i}$ is the classification predict by LDA or QDA.

For meta-modal, we compute the adjusted $R^2$ and the intercept of a simple linear model, as well as the smallest and largest absolute coefficients and their ratio beyond the simple linear model. $R^2$ measures the degree of fit of a model to data, with higher values indicating a better fit. Specifically, we calculate the adjusted $R^2$ of a linear model with interactions and a quadratic model with and without interactions, along with the ratio of the largest and smallest coefficients beyond the simple quadratic model without interactions.

Furthermore, features reflecting funnel structures are computed using nearest best/better clustering. These include the ratio of standard deviations and arithmetic mean based on the distances among the nearest neighbors and the nearest better neighbors, the correlation between distances of the nearest neighbors and the nearest better neighbors, the coefficient of variation of the distance ratios, and the correlation between fitness value and the count of observations to whom the current observation is the nearest better neighbor \cite{kerschke2015detecting}.

Information Content of Fitness Sequences (ICoFiS) approach \cite{munoz2014exploratory} is utilized to quantify the information content of a continuous landscape, such as smoothness, ruggedness, or neutrality. Specifically, we compute the maximum information content of the fitness sequence, settling sensitivity (which indicates the epsilon for which the sequence nearly consists of zeros only), epsilon-value, ratio of partial information sensitivity, and initial partial information.

Dispersion features are computed by comparing the dispersion of pairwise distances among the 'best' elements and the entire initial design, as well as the ratio and difference of the mean/median distances of the 'best' objectives versus 'all' objectives \cite{lunacek2006dispersion}.

Principal component analysis is employed to extract and analyze the main structure of the problems, including explained variance by convariance and correlation matrix of decision space, and proportion of variance explained by the first principal component. 

Linear model features are computed with the decision space being partitioned into a grid of cells. For each cell, the following features are calculated:
\begin{itemize}
    \item The length of the average coefficient vector.
    \item The arithmetic mean and standard deviation of the lengths of all coefficient vectors.
    \item The correlation among all coefficient vectors.
    \item The arithmetic mean and standard deviation of the ratios of the absolute maximum and minimum non-intercept coefficients within the cell.
    \item The max-by-min ratio of the standard deviations of the non-intercept coefficients.
    \item The arithmetic mean of the standard deviations of the non-intercept coefficients.
\end{itemize}

In addition to the above implicit characteristics, the solitary explicit characteristic of problem dimensionality is also encoded as one of the state variables.

\section{The proposed framework}
\label{rleag}

This section presents the proposed framework, rlDE. We begin by discussing the motivation behind the development of this methodology, followed by an overview of the framework and the procedural steps involved in operationalizing our theoretical constructs.

\subsection{Motivation}
BBOPs present unique challenges that defy a one-size-fits-all approach. The No Free Lunch theorem has prompted a shift towards more dynamic and intelligent optimization strategies for BBOPs. This quest for adaptability has led to the convergence of evolutionary computation and machine learning, particularly RL, which offers EAs the ability to learn and evolve in response to the optimization landscape. RL-assisted DE has emerged as a robust framework for exploring the solution space of BBOPs. However, existing studies have primarily focused on population characteristics, which raises several important questions.

Firstly, what is the relationship between population characteristics and problem characteristics? Can population characteristics be misleading? And is the cost of obtaining population characteristics lower than that of obtaining problem characteristics? To address these questions, we conducted a theoretical analysis.

On one hand, it is well established that the quality of feature approximation in ELA depends on the sample size \cite{renau2020exploratory}. A typical recommendation for the sample size $NS$ is around $50D$ when the functions are fast to evaluate \cite{kerschke2016low}. As for population size, Storn and Price suggested $NP\in [5D,10D]$ \cite{noman2008differential}, while Piotrowski suggested $NP=100$ for low-dimensional problems ($D \leq 30$) and $NP \in [3D,10D]$ for higher-dimensional problems \cite{piotrowski2017review}.  Given these recommendations, the sample size for population characteristics is generally smaller than that for problem characteristics. Consequently, population characteristics may potentially mislead the search direction, deviating from the true problem characteristics due to the smaller sample size, which can result in suboptimal solutions.

On the other hand, the cost of obtaining population characteristics increases with the number of iterations. If the fitness evaluations is fixed, defined as $maxFEs$, the iterations can be calculated as $\frac{maxFEs}{NP}$, where $NP$ represents the population size. Assuming that the number of used feature is $ftNum$, population characteristics-based DE computes each feature in each generation, leading to cost extra $ftNum \times \frac{maxFEs}{NP}$ evaluations. Problem characteristics-based DE only compute once before evolution, meaning that it cost only $ftNum$ extra evaluations. 

In light of these considerations, rlDE is proposed as an attempt to shift focus towards problem characteristics when automatically designing DEs. It leverages the decision-making capabilities of RL to meta-learn the mapping relationship between BBOPs and DEs offline, thereby realizing tailored DEs that are finely tuned to the problem characteristics. This approach transcends the limitations of manual algorithm design and the exhaustive search for optimal configurations, offering a more streamlined and efficient path to optimization.

By employing an enriched design space consisting of initialization strategies, mutation strategies, crossover strategies, and control parameters, rlDE is implemented via a DDQN to design matching DEs capable of effectively navigating the complex landscape of BBOPs.

\subsection{Overall framework}
\begin{figure*}[htbp]
  \centering
  \includegraphics[width=0.85\textwidth]{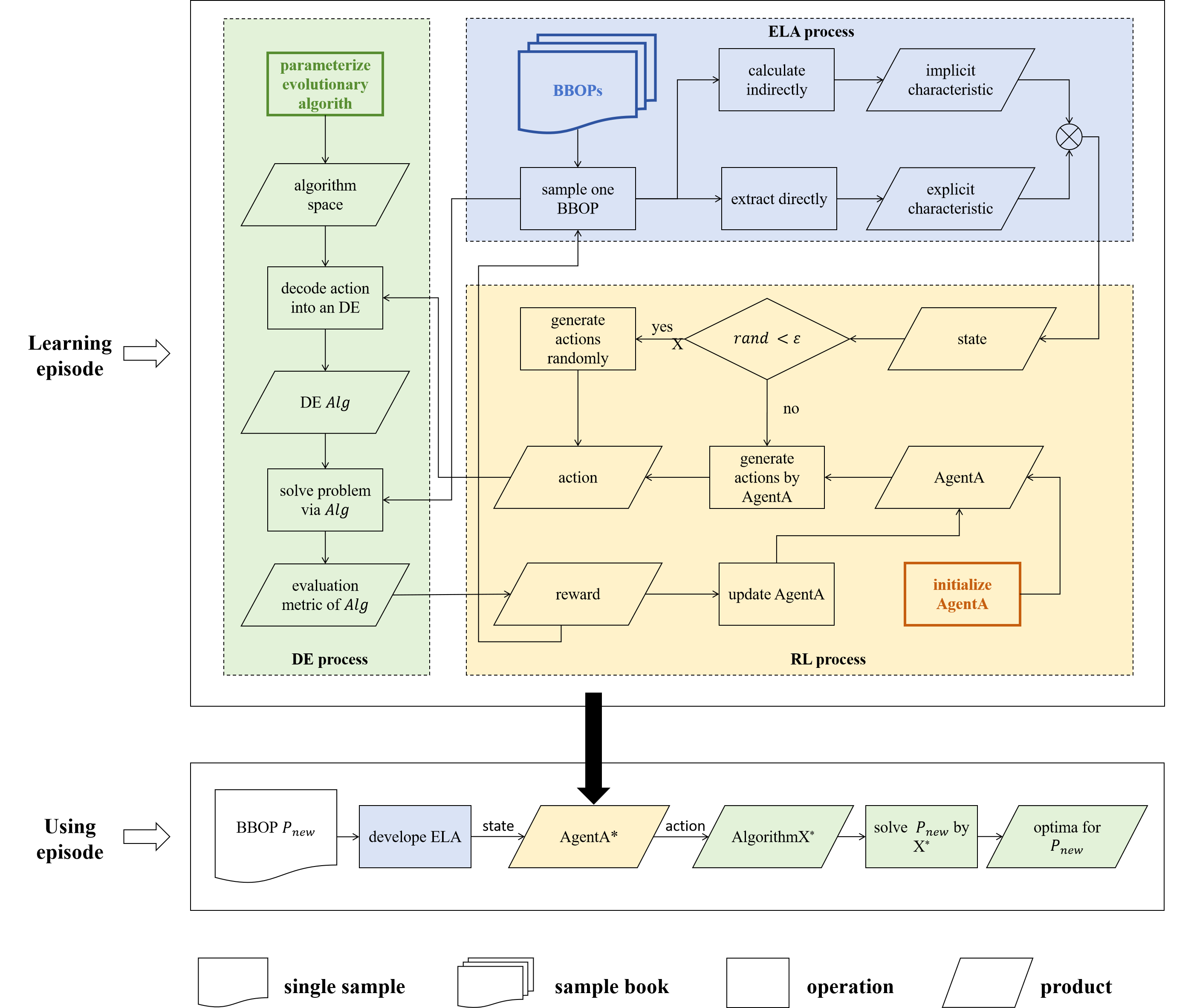}
  \caption{The overall framework of rlDE, encompassing three parts: ELA, RL process and DE process, where the starting points of each part are emphasized with different colors.}
  \label{figoverallframework}
\end{figure*}

The framework of rlDE illustrated in the \fref{figoverallframework}, integrates ELA with both RL and DE processes to provide a comprehensive approach to solve BBOPs. This framework leverages the strengths of each component to navigate complex problem spaces effectively. ELA provides a multidimensional understanding of problem characteristics, which is crucial for informing the sampling of a problem before it is subjected to black-box optimization. The RL process learns from experiences and refines its decision-making capabilities in the learning episode through iterative training. The DE process provides feedback to the RL process.

To realize the automatic design of DE tailored to a specific BBOP, the learning episode focuses on training an RL agent to discover the mapping between DE and BBOP before the using episode. In this episode, ELA is performed on a considerable number of BBOPs to serve as the input for the RL agent, denoted as $AgentA$. It designs various DEs based on the defined design space and is updated by the reward obtained from executing the DE process. Ultimately, this episode outputs an optimal RL agent, denoted as $AgengtA^*$, for use in the subsequent episode.

\subsection{Procedural steps}
\subsubsection{Learning episode}
The learning episode involves three parts: ELA process, RL process, and DE process, with the beginning of collecting BBOPs, initializing an RL agent and parameterizing DE separately. 

First, we need to collect enough BBOPs serving as training and testing sets.
Second, an RL agent is initialized by random initialization, zero-value initialization, priori knowledge-based initialization, or heuristic initialization, etc. For wider generality, random initialization is utilized in our implementation.  
Third, parameterization in DE process defines the design space, as the output actions of the agent. In our implementation, the selection paradigm of DE is limited to the $\mu+\lambda$ template, where $\mu$ denotes the number of target vectors and $\lambda (\geq \mu)$ represents the size of trial vectors. This template is detailed in Algorithm \ref{mulambdaTemplate}, which populates the next generation with the best $\mu$ vectors from the combined target and trial vectors.

\begin{algorithm} 
  \caption{$\mu+\lambda$ paradigm of DE}
  \label{mulambdaTemplate} 
  \begin{algorithmic}[1] 
    \State Initialize population $P$ with $NP$ individuals
    \State Evaluate fitness of each individual in $P$ 
    \While{termination condition not met}
      \State select $\mu$ target vectors from $P$
      \State generate $\lambda$ trail vectors based on $\mu$ target vectors
      \State select $\mu$ vectors as new $P$ from ($\mu$ target vectors, $\lambda$ trail vectors)
    \EndWhile 
  \end{algorithmic} 
\end{algorithm}

The design space consists of configurable components and control parameters. For configurable components, three types are considered: initialization strategy, mutation strategy, and crossover strategy. Specifically, the initialization strategy includes random, Latin hypercube sampling (LHS), uniform, normal, and tent mapping (TM) strategies. The mutation strategy is denoted by DE/x/y, where x represents the base vector to be perturbed and y is the number of difference vectors. The crossover strategy comprises binomial and exponential strategies \cite{deb2000efficient}. For control parameters, population size $NP$, scale factor $F$, and crossover rate $Cr$ are considered.

The detailed options in the design space are summarized in Table \ref{tabDesignSpace}, where $lb$ and $ub$ represent the lower and upper bounds of the problem, and $np.random.*$ represents functions from the Python package "numpy.random" \cite{harris2020array} among initialization strategies. In the mutation strategy, $X_{i}$, $V_{i}$, $U_i$ represent the $i$-th target vector, donor vector, and trial vector, respectively, while $x_{i,j}$, $v_{i,j}$, $u_{i,j}$ represent the $j$-th element in the corresponding vector, and $r1$ to $r5$ are different integers randomly generated between 1 and $NP$. For the crossover strategy, $rand_{i,j}$ is a uniformly distributed random number, $j_{rand}$ is a randomly chosen index, $k$ is a random integer representing the starting position of the crossover, and $L$ is the random length of the crossover.

\begin{algorithm} 
  \caption{Pseudocode of LHS}
  \label{algLHS} 
  \begin{algorithmic}[1] 
    \State \textbf{Input:} $NP$: number of samples, $D$: number of dimensions, $lb$: lower bound vector, $ub$: upper bound vector
    \State \textbf{Output:} $P$: $NP \times D$ matrix of initialized sample points
    \For {$d \in [1,D]$}
      \State Divide $NP$ equal-length intervals between $lb_d$ and $ub_d$
      \State Select one point from each interval randomly
      \State Combine these points into an $NP \times 1$ vector as the $d$-th sample points
    \EndFor
    \State Combine $D$ vectors into a $NP \times D$ matrix as $P$
  \end{algorithmic} 
\end{algorithm}

\begin{algorithm} 
  \caption{Pseudocode of Tent Mapping Initialization}
  \label{algTM} 
  \begin{algorithmic}[1] 

    \State \textbf{Input:} $NP$: number of samples, $D$: number of dimensions, $lb$: lower bound vector, $ub$: upper bound vector
    \State \textbf{Output:} $P$: $NP \times dim$ matrix of initialized sample points
    \State $alpha \leftarrow 0.7$
    \State $T \leftarrow np.random.rand(NP,D)$
    \For {$i \in [1, NP]$}
      \For {$j \in [1, dim]$}
        \If {$T[i, j] < alpha$}
          \State $T[i, j] \leftarrow \frac{T[i, j]}{alpha}$
        \Else
          \State $T[i, j] \leftarrow \frac{1 - T[i, j]}{1 - alpha}$
        \EndIf
      \EndFor
    \EndFor
    \State $P\leftarrow lb + T \cdot (ub - lb)$
    \State \textbf{return} $P$
  \end{algorithmic} 
\end{algorithm}

\begin{table*}[htbp]
\centering
\caption{Design space of DE in rlDE}
  \begin{tabular}{ccc}
    \toprule
    \textbf{Design object} & \textbf{Options} & \textbf{Detail} \\
    \midrule
    \multirow{5}{*}{Initialization strategy} &random & $P=np.random.rand(NP,D)*(ub-lb)+lb$ \\
    & latin hypercube sampling & $P=LHS(NP,D,lb,ub)$, see in \alref{algLHS} \\
    & uniform & $P=np.random.uniform(low=lb,high=ub, size=(NP,D))$ \\
    & normal & $P=np.random.rand(NP,D)*((ub-lb)/6)+(ub+lb)/2$ \\
    & tent mapping &$P=TM(NP,D,lb,ub)$, see in \alref{algTM} \\
    \hline
    \multirow{10}{*}{Mutation strategy} & DE/rand/1 & $V_{i} = X_{r1} + F*(X_{r2} - X_{r3})$ \\
    & DE/rand/2 & $V_{i}  = X_{r1} + F*(X_{r2} - X_{r3}) + F*(X_{r4} - X_{r5})$ \\
    & DE/best/1& $V_{i}  = X_{best} + F*(X_{r1} - X_{r2})$ \\
    & DE/best/2& $V_{i}  = X_{best} + F*(X_{r1} - X_{r2}) + F*(X_{r3} - X_{r4})$ \\
    & DE/current-to-rand/1&$ V_{i} = X_{i} + F*(X_{r1} - X_{i}) + F*(X_{r2} - X_{r3})$ \\
    & DE/current-to-best/1&$ V_{i} = X_{i} + F*(X_{best} - X_{i}) + F*(X_{r1} - X_{r2})$ \\ 
    & DE/rand-to-best/1& $ V_{i} = X_{r1} + F*(X_{best} - X_{r2}) + F*(X_{r3} - X_{r4})$ \\ 
    & DE/current-to-rand/2&$ V_{i} = X_{i} + F*(X_{r1} - X_{i}) + F*(X_{r2} - X_{r3}) + F*(X_{r4} - X_{r5})$ \\
    & DE/current-to-best/2& $ V_{i} = X_{i} + F*(X_{best} - X_{i}) + F*(X_{r1} - X_{r2}) + F*(X_{r3} - X_{r4})$ \\ 
    & DE/rand-to-best/2& $ V_{i} = X_{r1} + F*(X_{best} - X_{r1}) + F*(X_{r2} - X_{r3})  + F*(X_{r4} - X_{r5})$ \\ 
    \hline 
    \multirow{2}{*}{Crossover strategy} & binomial & $u_{i,j}=\begin{cases} 
      v_{i,j} & \text{if } (rand_{i,j} \leq Cr || j=j_{rand}) \\
      x_{i,j} & \text{otherwise}.
      \end{cases}$ \\
    & exponential & $u_{i,j}=\begin{cases}
      v_{i,j} & \text{for} j=k,\ldots,k-L+1 \in [1, D], \\
      x_{i,j} & \text{otherwise}.
      \end{cases}$\\
    Population size  & discrete & $NP \in \{5D,7D,9D,11D,13D\}$\\
    Scale factor& discrete & $\{ F \in \mathbb{R} \mid F = k \cdot 0.05, \, k \in \mathbb{Z}, \, 0 \leq k \leq 40 \}$ \\
    Crossover rate & discrete & $\{ Cr \in \mathbb{R} \mid Cr = k \cdot 0.1, \, k \in \mathbb{Z}, \, 0 \leq k \leq 20 \}$ \\
    \bottomrule
  \end{tabular}
  \label{tabDesignSpace}
\end{table*}

\begin{figure} [htbp]
  \centering
  \includegraphics[width=0.45\textwidth]{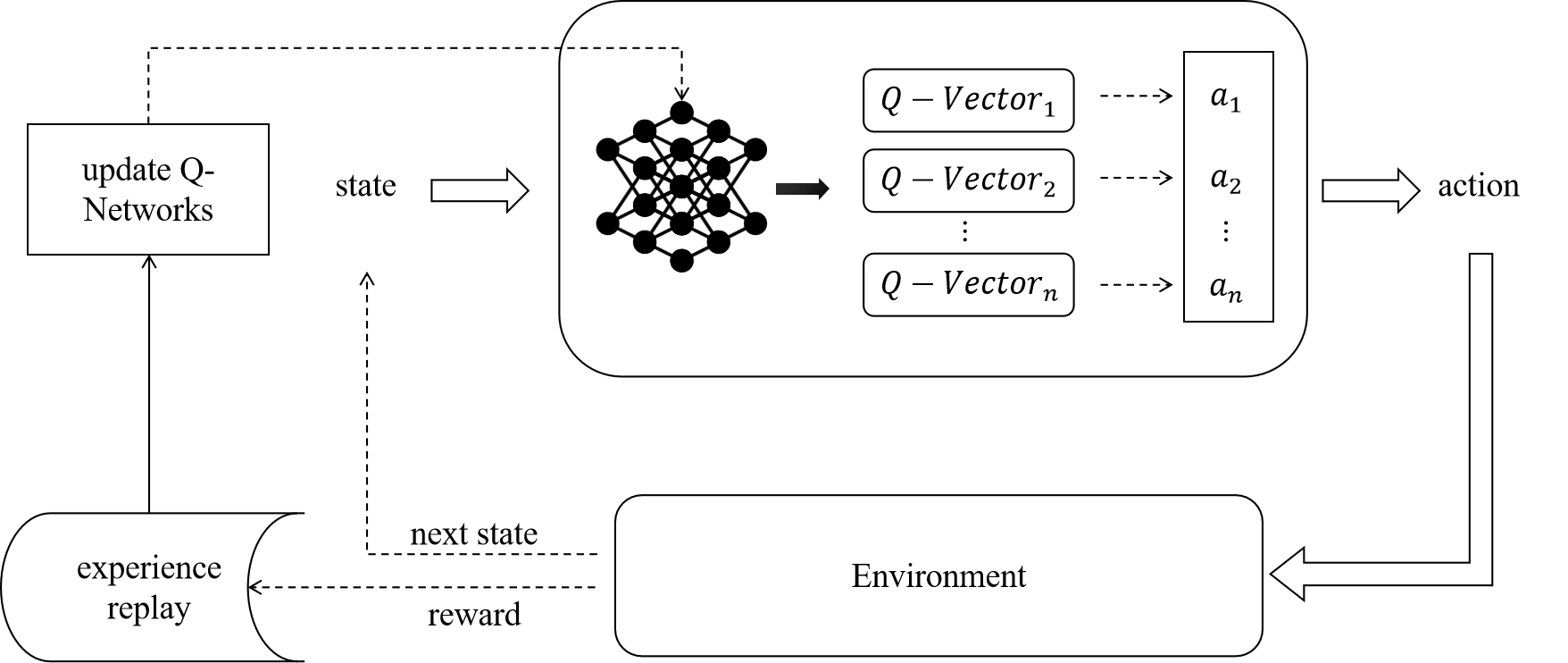}
  \caption{MADQN process}
  \label{madqnProcess}
\end{figure}

\textbf{ELA process} conducts a comprehensive assessment of the problem's characteristics as the input state of the agent, encompassing both explicit and implicit features, as prementioned in \sref{lowfeatures}. 
\textbf{RL process} is dominating the learning episode, where DDQN is employed to learn the optimal actions in a given state by interacting with the environment and receiving feedback in the form of rewards or penalties. In order to decide multiple types of strategies simultaneously, the principle of Multi-Armed Bandit(MAB) \cite{slivkins2019introduction} is introduced into DDQN, which is termed MADQN. The process of MADQN is depicted in \fref{madqnProcess}. In MADQN, the implicit and explicit characteristics are integrated and transferred as the specific state $s$. The action $a$ represents the choice of $n$ types of design objects. A deep neural network is used to approximate the $Q$ function, i.e. $Q(s, a) \approx Q(s,a,\theta)$, where $\theta$ is the parameter of the neural network and initialized by random strategy. The input of the network is state while the output of the network is a matrix made up of $n$ Q-value vectors corresponding to $n$ actions. The network will be updated through an experience replay buffer $B$, where stores the evaluation of the current state and the generation of new actions. Specifically, given a period of time step $T$, $T$ records from $B$ will be sampled, and $loss$ will be calculated according to \eref{calloss}, where $\{B_{i1}, B_{i2}, B_{i3}, B_{i4}\}$ is the $i$-th record from the sampled records which represents respectively the input state, the output action, corresponding reward and the next state at some time stamp. The corresponding reward can be calculated after the DE process. Adam optimizer is used then to minimize $loss$ to update $\theta$ of the network.

\begin{figure*}
  \centering
  \begin{equation}
    \label{calloss}
    loss = \frac{\sum_{i=1}^T B_{i3} + \gamma * \hat{Q}(B_{i4}, argmax_a Q(B_{i4},a|\theta)| \hat{\theta}) - Q(B_{i1}, B_{i2}|\theta)}{T}
  \end{equation}
\end{figure*}

To better balance the exploration and exploitation when training RL agent, $\epsilon$-greedy strategy is introduced as well. This iterative process allows RL agent to learn from past experiences and refine its decision-making capabilities. 

\textbf{DE process} executes each $AlgorithmX$ decoded according to $a_t$ on the sampled BBOP, and the evaluation metric on $AlgorithmX$ will be converted to $r_t$. In this implementation, each BBOP is regarded as a minimization problem, and the evaluation metric is set to the solution quality, i.e. obtaining minimum objective values, denoted as $f_t^*$. It is related to the design of reward $r_t$.

Actually, the design of the reward function is also a crucial task in RL and can take various forms \cite{hu2020learning}. In this context, it is formulated as the \eref{rew}. Further research on the design of the reward function will be conducted in subsequent work.

\begin{equation}
  r_t = \mathrm{e}^{-f_t^*}
  \label{rew}
\end{equation}

In summary, the algorithm of training episode of rlDE can be described in \alref{algTraining}.

\begin{algorithm} 
  \caption{training episode of rlDE}
  \label{algTraining} 
  \begin{algorithmic}[1] 
    \State Initialize Q-network $Q$ with parameter $\theta$ and target Q-network $\hat{Q}$ with parameter $\hat{\theta}$ randomly
    \State Parameterize DE
    \State Instantiate $M$ BBOPs
    \State Initialize control variable $t=1$, constant time step $T$, constant target Q-network update frequency $\hat{T}$
    \While{termination condition of training not met}
      \State Sample BBOP $p_t$ among $M$ BBOPs randomly
      \State Calculate state $s_t$ via ELA on $p_t$
      \While{$t\leq M$}
        \State set $count = 1$
        \If{$rand < \epsilon$}
          \State Generate action $a_t$ randomly
        \Else
          \State Generate action $a_t$ by Q-network where $a_t = argmax_a Q(s_t, a|\theta)$
        \EndIf
        \State Generate Algorithm $Alg_t$ according to action $a_t$
        \State Evaluate performance $performance_t$ of $Alg_t$ solving $p_t$
        \State Calculate reward $r_t = f(performance_t)$
        \If{$t=M$}
          \State $next = 1$
        \Else 
          \State $next = t+1$
        \EndIf
        \State Calculate state $s_{next}$
        \State Store the record 
        \If{the number of records in $B$ is larger than $T$}
          \State Sample $T$ records from $B$
          \State Calculate $loss$
          \State Update $\theta$ by Adam optimizer based on $loss$
        \EndIf
        \If{$count \% \hat{T} == 0$}
          \State $\hat{\theta} = \theta$
        \EndIf
        \State $count = count + 1$
        \State $t=next$
      \EndWhile
    \EndWhile 
  \end{algorithmic} 
\end{algorithm}

\subsubsection{Using episode}
In the using episode, $AgentA^*$ will be used to deal with the new BBOP $P_{new}$. ELA on the $P_{new}$ is developed, $AlgorithmA^*$ is designed by $AgentA^*$, and the optima of $P_{new}$ denotes output after executing $AlgorithmA^*$ on $P_{new}$. 

\section{Experiments}
\label{exp}

The framework of rlDE is evaluated using a synthetic test suite real-parameter Black-Box Optimization Benchmarking 2009 (BBOB2009) \cite{hansen2009real}. BBOB2009 consists of 24 problems, all of which can be instantiated numerical instances. The MetaBox platform \cite{ma2024metabox} is utilized for this evaluation, which includes several built-in algorithms for comparison. This section provides details on the experimental setup, results, and analysis.

\subsection{Experimental setup}
All BBOB2009 problems are set to a dimension of 10, which are divided into a learning set and a testing set. In the learning episode, 75\% of the problem instances from the noiseless BBOB2009 are used following the setup in \cite{ma2024metabox}, which can enhance the usability and flexibility. Our DDQN architecture includes five layers: an input layer with 62 neurons representing 62 characteristic values, three fully connected hidden layers, and an output layer with 5 neurons that can be decoded into specific configurations. The maximum learning steps at the meta-level are set to 10,000, and the maximum evaluations at the base-level are set to  $2000D$. According to \cite{ma2024metabox}, 18 training functions are predetermined, while the remaining six noiseless functions, including $f_1, f_5, f_6, f_{10}, f_{15}$, and $f_{20}$, are used for testing. Each combination of a testing function and a test algorithm is run 31 times. 

The performance of rlDE is compared with four state-of-the-art RL-assisted DEs and four traditional DEs. Specifically, RL-assisted DEs consist of DEDDQN \cite{sharma2019deep}, DEDQN \cite{tan2021differential}, LDE \cite{sun2021learning} and RL\_HPSDE \cite{TAN2022101194}. Traditional DEs consist of classical DE/rand/1, JDE21 \cite{brest2021self}, MadDE \cite{biswas2021improving}, and NL\_SHADE\_LBC \cite{stanovov2022nl}. In addition. Random Search (RS) is included as a baseline,  which randomly samples candidate solutions from the search space. The parameter values for each compared algorithm are adopted according to the recommendations in the respective cited papers. These algorithms are evaluated based on the best objective value with an accuracy of 1e-8 and a fixed maximum number of evaluations.

Statistical analysis is conducted using the Wilcoxon rank-sum two-sided test to assess the significance of differences between the results obtained by the comparative algorithms and rlDE at the 95\% confidence level. 


\subsection{Experimental results and analysis}
\begin{figure*}[htbp]
  \centering
  \includegraphics[width= 0.9\textwidth]{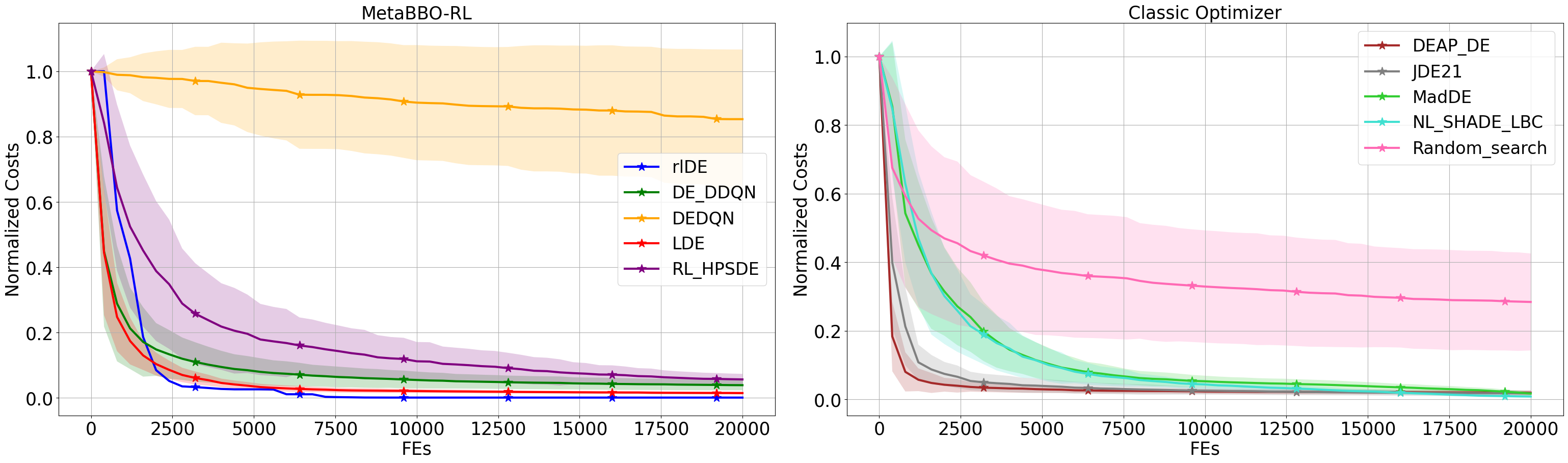} 
  \caption{Convergence curve of normalized averaged best $f^*$ as evolving of all algorithms performing on all testing functions, where X-axis represents the costed fitness evaluations and the Y-axis represents the normalized average best objective.}
  \label{figRes}
\end{figure*}

This section provides the results and analysis of experiments on BBOB2009. \fref{figRes} presents the natural logarithm of the average minimum objective values found by each algorithm across BBOB2009, with increasing evaluations conducted over 31 runs. Due to spatial constraints, the results are displayed separately. The left panel covers RL-assisted DEs, while the right panel covers traditional DEs. The convergence curves across various test suites indicate that rlDE, depicted by the dark blue line, exhibits a relatively swift convergence during the initial phases of function evaluations and ultimately attains the optimal objective value. Among the RL-assisted DEs, rlDE demonstrates superior performance, followed by LDE, DEDDQN, RL\_HPSDE, and DEDQN. The traditional DEs exhibit similar levels of performance, as observed in the right panel of \fref{figRes}, with the classical DE achieving the fastest convergence. When comparing these two segments, the robustness of rlDE is apparent, given its consistent achievement of optimal objectives under diverse test conditions. This consistent performance across multiple graphs suggests that rlDE is capable of effectively addressing a wide range of optimization problems, underscoring its adaptability and resilience. 

\begin{table*}[!ht] \footnotesize
  \centering
  \caption{Results obtained by rlDE on BBOB2009 testing functions}
  \begin{tabular}{cccccc}
    \toprule
        Problem & Worst & Best & Median & Mean & Std \\  \midrule
        $f_{1}$ & 8.485e-009 & 8.485e-009 & 8.485e-009 & 8.485e-009 & 0.e+000 \\ 
        $f_{5}$ & 0.e+000 & 0.e+000 & 0.e+000 & 0.e+000 & 0.e+000  \\ 
        $f_{6}$ & 1.042e-007 & 1.042e-007 & 1.042e-007 & 1.042e-007 & 0.e+000 \\
        $f_{10}$ & 9.316e-009 & 9.316e-009 & 9.316e-009 & 9.316e-009 & 0.e+000  \\
        $f_{15}$ & 9.95e-001 & 9.95e-001 & 9.95e-001 & 9.95e-001 & 0.e+000 \\
        $f_{20}$ & 1.224e+000 & 1.224e+000 & 1.224e+000 & 1.224e+000 & 0.e+000 \\
    \bottomrule
  \end{tabular}
  \label{tabresrldenoiseless}
\end{table*}

\tref{tabresrldenoiseless} presents the worst, best, median, mean and standard deviation values oobtained from 31 runs of rlDE on each BBOB2009 testing function. Each standard deviation is zero, indicating that all runs yielded identical results. This lack of variability suggests that the algorithm is either highly stable or has converged to a single optimal solution for each function, demonstrating that rlDE is consistently effective in finding solutions for these testing functions.

\begin{figure*}
  \centering
  \subfloat[curve on $f_1$]{\includegraphics[width = 0.3\textwidth]{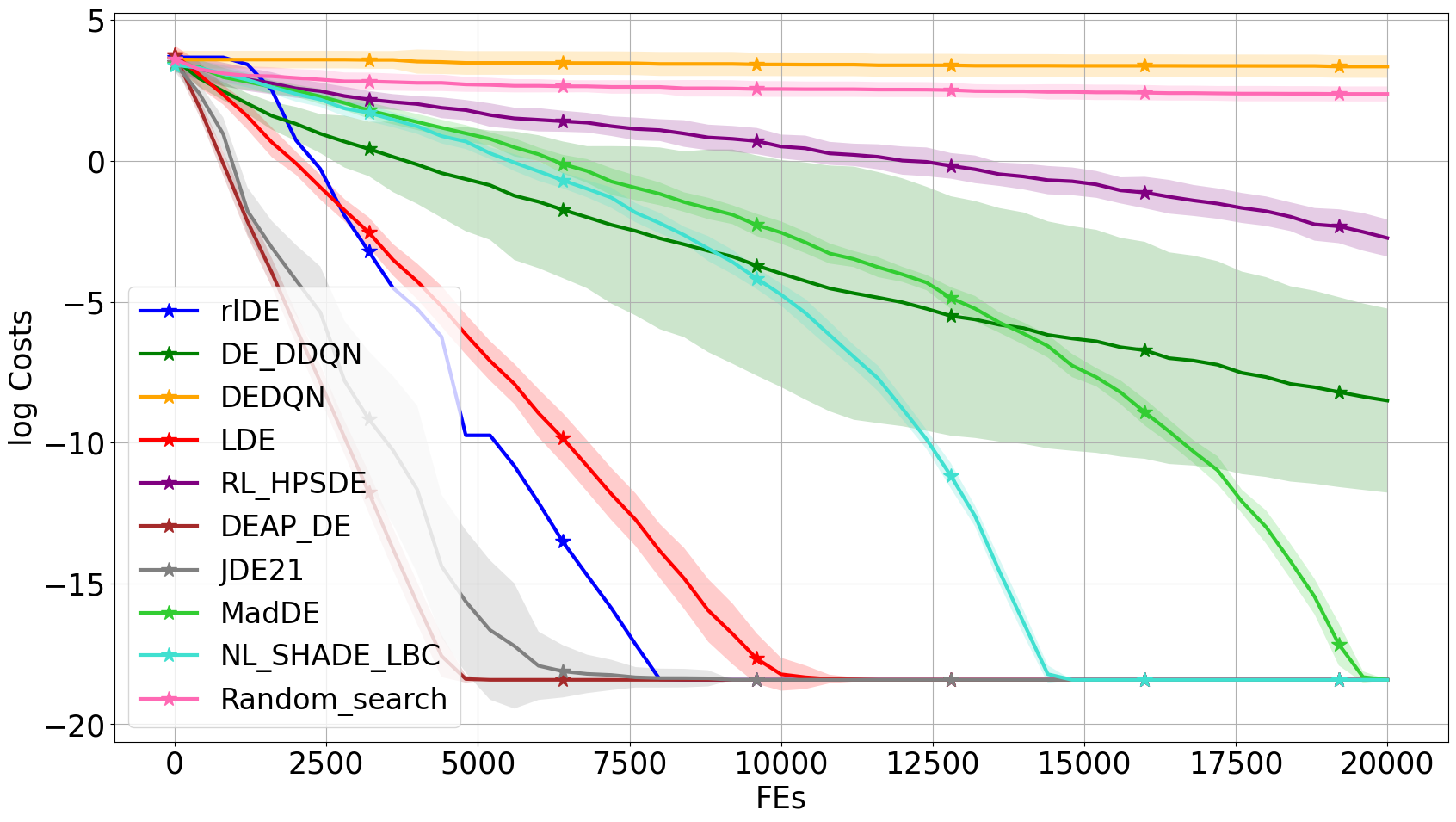}}
  \subfloat[curve on $f_5$]{\includegraphics[width = 0.3\textwidth]{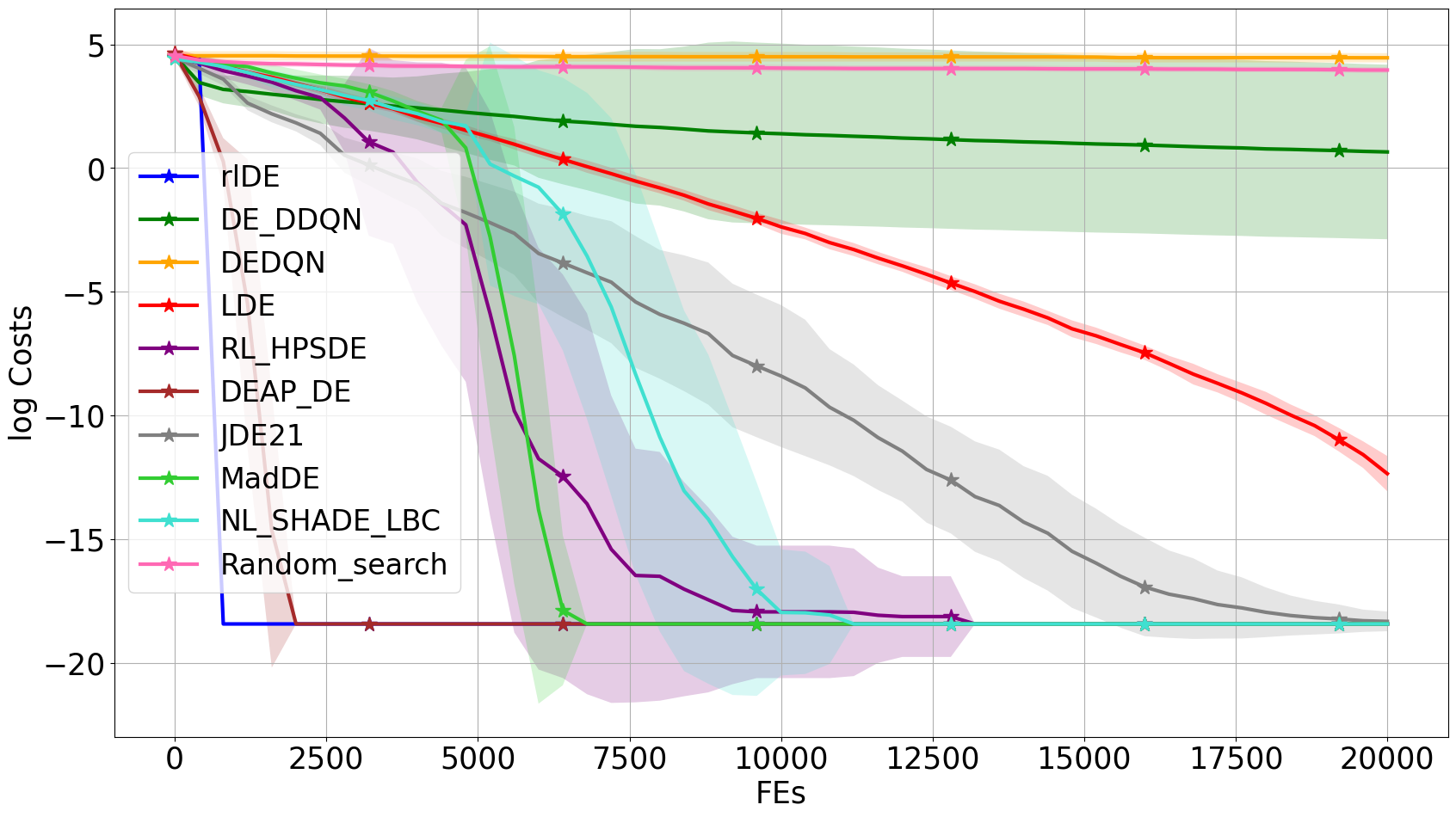}}
  \subfloat[curve on $f_6$]{\includegraphics[width = 0.3\textwidth]{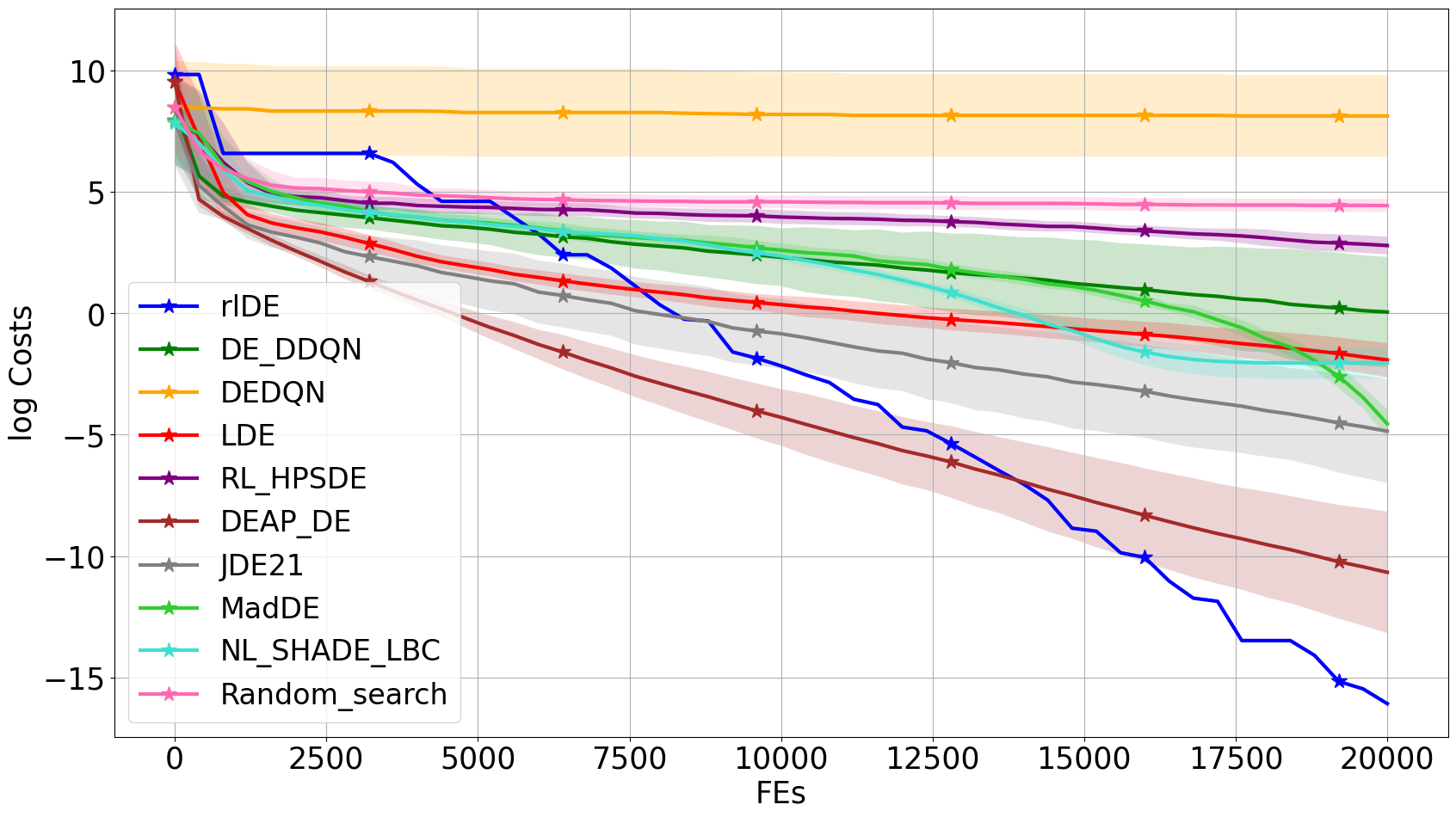}}\\
  \subfloat[curve on $f_{10}$]{\includegraphics[width = 0.3\textwidth]{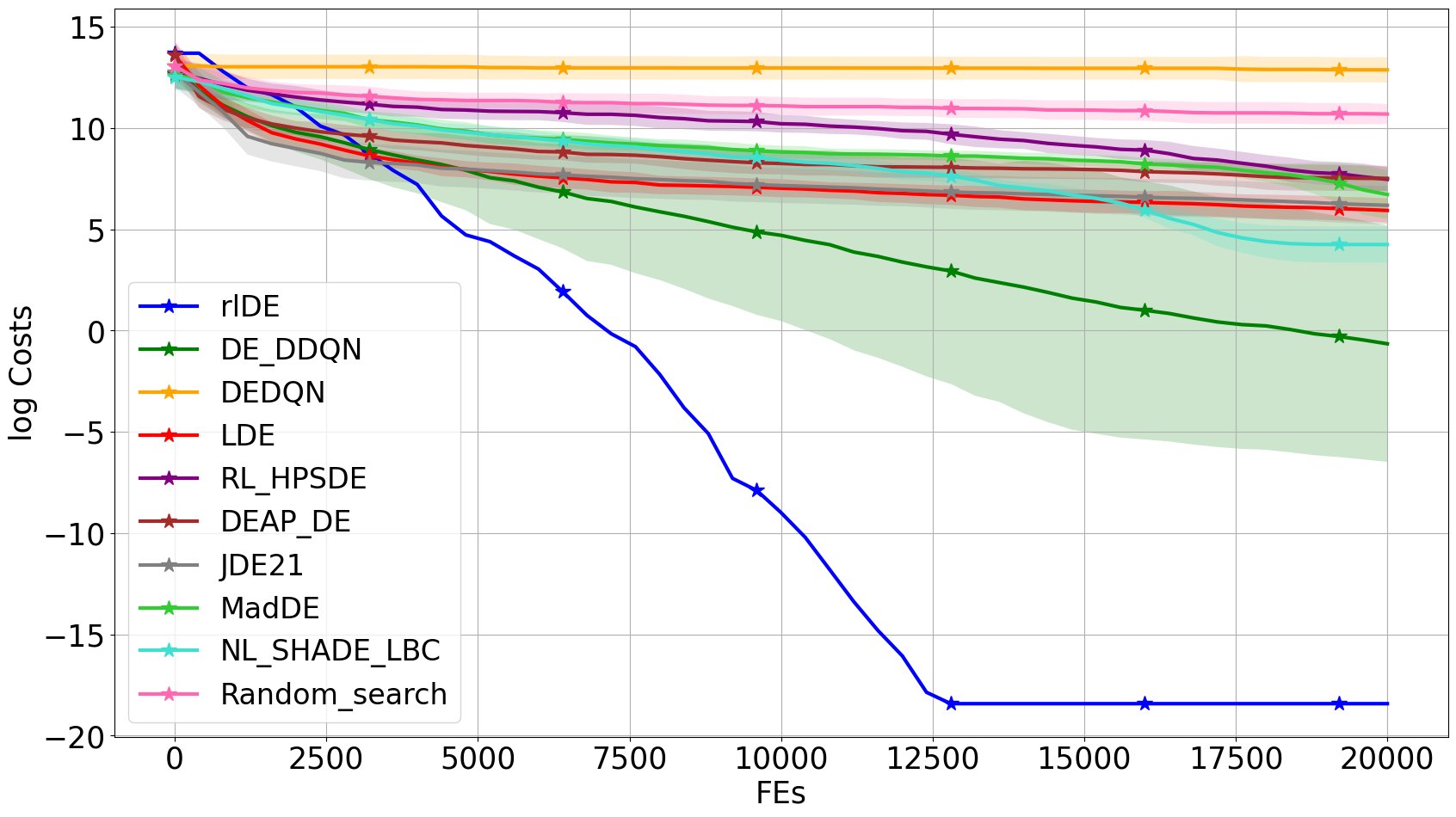}}
  \subfloat[curve on $f_{15}$]{\includegraphics[width = 0.3\textwidth]{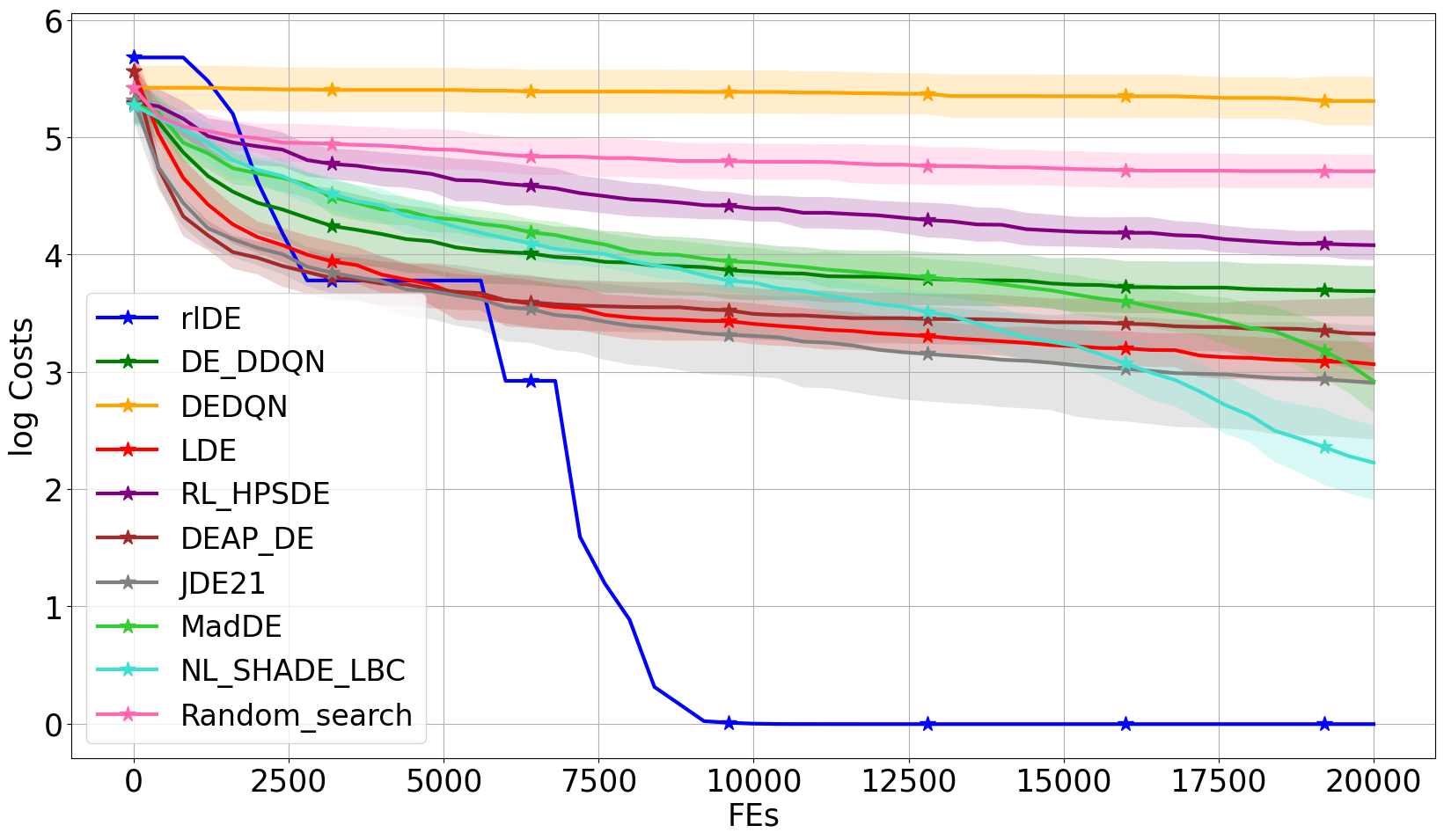}}
  \subfloat[curve on $f_{20}$]{\includegraphics[width = 0.3\textwidth]{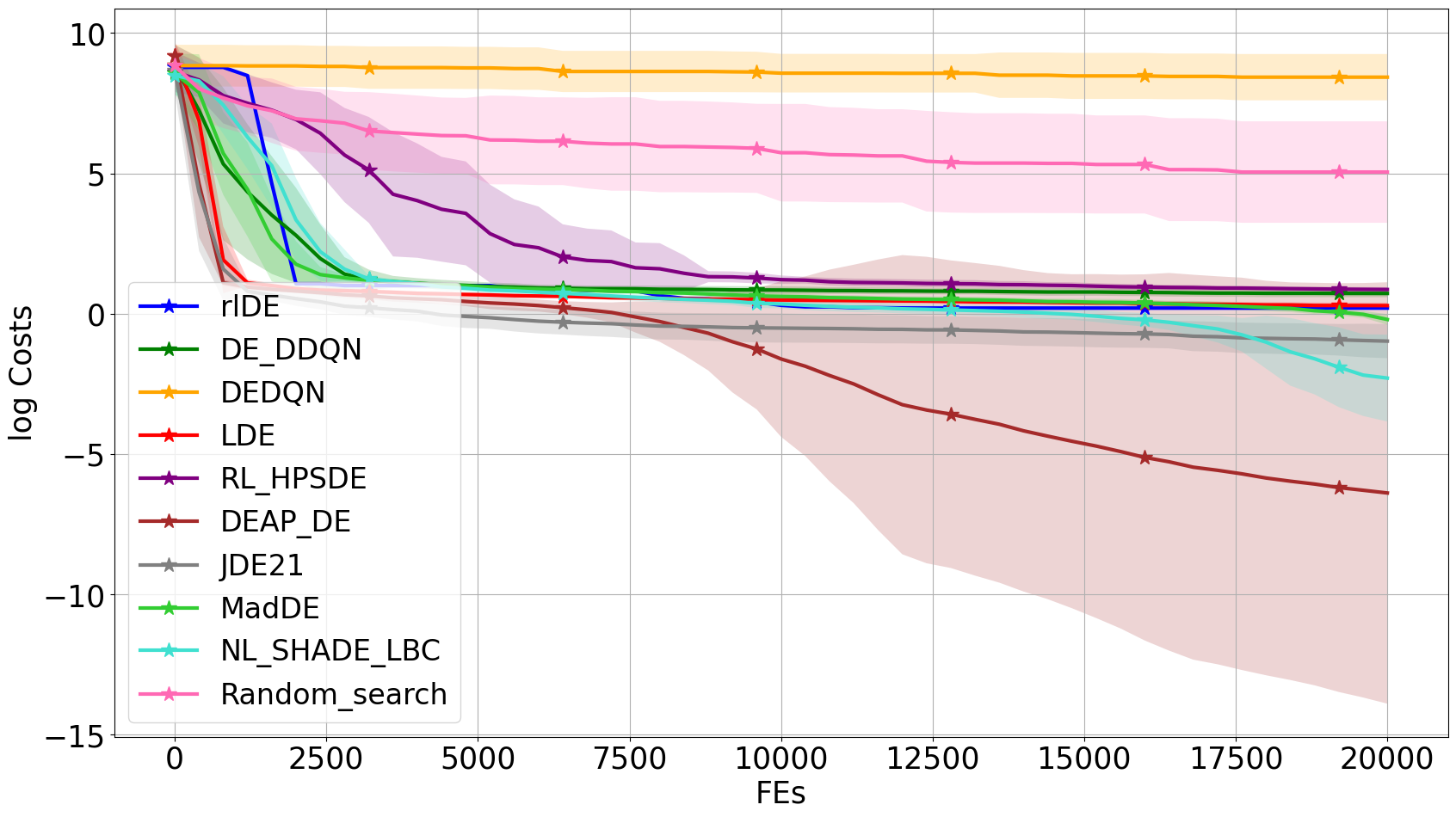}}
  \caption{Per-instance curve of logarithm best objective with increasing costed evaluations}
  \label{figPer}
\end{figure*}

\tref{tabOverallPerf} presents the detailed average obtained objective value $v_{avg}$ with standard deviation $v_{std}$ in the bracket of per-instance , where the symbols "-","+", and "=" separately indicate whether the given algorithm performs significantly worse, significantly better, or equal to rlDE. \fref{figPer} llustrates the logarithmic best objective curves for each function. It can be observed that rlDE exhibits competitive performance across all functions, achieving top results on $f_5$, $f_6$, $f_{10}$ and $f_{20}$. For $f_1$, DE, JDE21 and NL\_SHADE\_LBC performs better than rlDE, LDE and MadDE perform on par with to rlDE, while all of the other RL-assisted DEs perform worst than rlDE.  For $f_{20}$, comparative RL-assisted DEs falls behind rlDE while the traditional DEs perform better than rlDE. Comparing each algorithm with rlDE rom the view of the row in the \tref{tabOverallPerf}, rlDE outperforms all the RL-assisted DEs on these testing functions.

\begin{table*}[htbp] \tiny
  \centering
  \caption{Per-instance optimization results $v_{avg}(v_{std})$ of each algorithm on BBOB2009, where the symbols "-","+", and "=" separately indicate whether the given algorithm performs significantly worse, significantly better, or equal to rlDE.}
  \begin{tabular}{ccccccc}
    \toprule
    Algorithm & \multicolumn{1}{c}{$f_1$} & \multicolumn{1}{c}{$f_5$} & \multicolumn{1}{c}{$f_6$} & \multicolumn{1}{c}{$f_{10}$} & \multicolumn{1}{c}{$f_{15}$} & \multicolumn{1}{c}{$f_{20}$} \\
    \midrule
    DE\_DDQN &  5.714e-4(2.991e-4)-  &  4.12e+0(2.710e+0)-  &  1.581e+0(4.866e-1)-  &  3.518e+0(2.236e+0)-  &  4.081e+1(8.284e+0)-  &  2.081e+0(2.635e-1)-   \\
    DEDQN  &  3.046e+1(9.843e+0)-  &  8.777e+1(1.571e+1)-  &  8.809e+3(9.91e+3)-  &  4.554e+5(2.242e+5)-  &  2.062e+2(4.009e+1)-  &  5.810e+3(3.135e+3)-   \\
    LDE  &  8.488e-9(1.386e-9)=  &  5.603e-6(4.335e-6)-  &  1.767e-1(9.327e-2)-  &  4.478e+2(2.713e+2)-  &  2.178e+1(3.838e+0)-  &  1.345e+0(1.383e-1)-   \\
    RL\_HPSDE  &  7.858e-2(4.599e-2)-  &  0.e+0(0.e+0)=  &  1.735e+1(6.216e+0)-  &  2.102e+3(1.527e+3)-  &  5.956e+1(7.388e+0)-  &  2.381e+0(2.023e-1)-   \\
    DEAP\_DE &  7.593e-9(1.714e-9)+  &  0.e+0(0.e+0)=  &  1.253e-4(1.755e-4)-  &  2.167e+3(1.170e+3)-  &  2.883e+1(6.641e+0)-  &  1.640e-1(1.542e-1)+   \\
    JDE21 &  5.1e-9(2.224e-9)+  &  1.207e-8(1.387e-8)-  &  5.636e-2(1.519e-1)-  &  7.127e+2(5.921e+2)-  &  2.033e+1(8.560e+0)-  &  4.414e-1(2.298e-1)+   \\
    MadDE &  8.238e-9(1.285e-9)=  &  0.e+0(0.e+0)=  &  1.260e-2(8.306e-3)-  &  1.171e+3(6.41e+2)-  &  1.920e+1(4.914e+0)-  &  8.296e-1(1.487e-1)+   \\
    NL\_SHADE\_LBC &  7.658e-9(1.555e-9)+  &  0.e+0(0.e+0)=  &  1.555e-1(9.075e-2)-  &  1.051e+2(1.115e+2)-  &  9.722e+0(3.063e+0)-  &  2.040e-1(1.528e-1)+   \\
   RS &  1.111e+1(2.588e+0)-  &  5.295e+1(5.920e+0)-  &  8.706e+1(2.236e+1)-  &  4.881e+4(2.044e+4)-  &  1.121e+2(1.502e+1)-  &  3.669e+2(3.180e+2)-  \\ \hline
    sum(-/+/=)	& 4/3/2 &	5/0/4 &	9/0/0	& 9/0/0	& 9/0/0	& 5/4/0 \\
\bottomrule
    \end{tabular}%
  \label{tabOverallPerf}%
\end{table*}%

Besides, a simple-calculation but comprehensive indicator Aggregated Evaluation Indicator (AEI) proposed by Ma et al. \cite{ma2024metabox} is employed to provide a holistic view of the performance. AEI combines three evaluation metrics including best objective value, costed evaluations to achieve the pre-defined accuracy 1e-8, and runtime complexity. Those metrics of each run $n$ on problem $k$ are denoted as $v_{obj}^{k,n}$, $v_{fes}^{k,n}$, and $v_{time}^{k,n}$ separately. AEI can be calculated as follows:
\begin{equation}
AEI = \frac{1}{K} \sum_{k=1}^K e^{Z_{obj}^k+Z_{fes}^k+Z_{time}^k}
\end{equation}
where $K$ represents all testing functions, and $Z_{*}^k$ is calculated via Z-score normalization of results obtained in repeated runtimes, according to \eref{zscore}.

\begin{equation}
  Z_{*}^k = \frac{1}{N} \sum_{n=1}^N \frac{v_*^{k,n}}{\sigma_{*}}
  \label{zscore}
\end{equation}

\begin{figure}
  \centering
  \includegraphics[width = 0.45\textwidth]{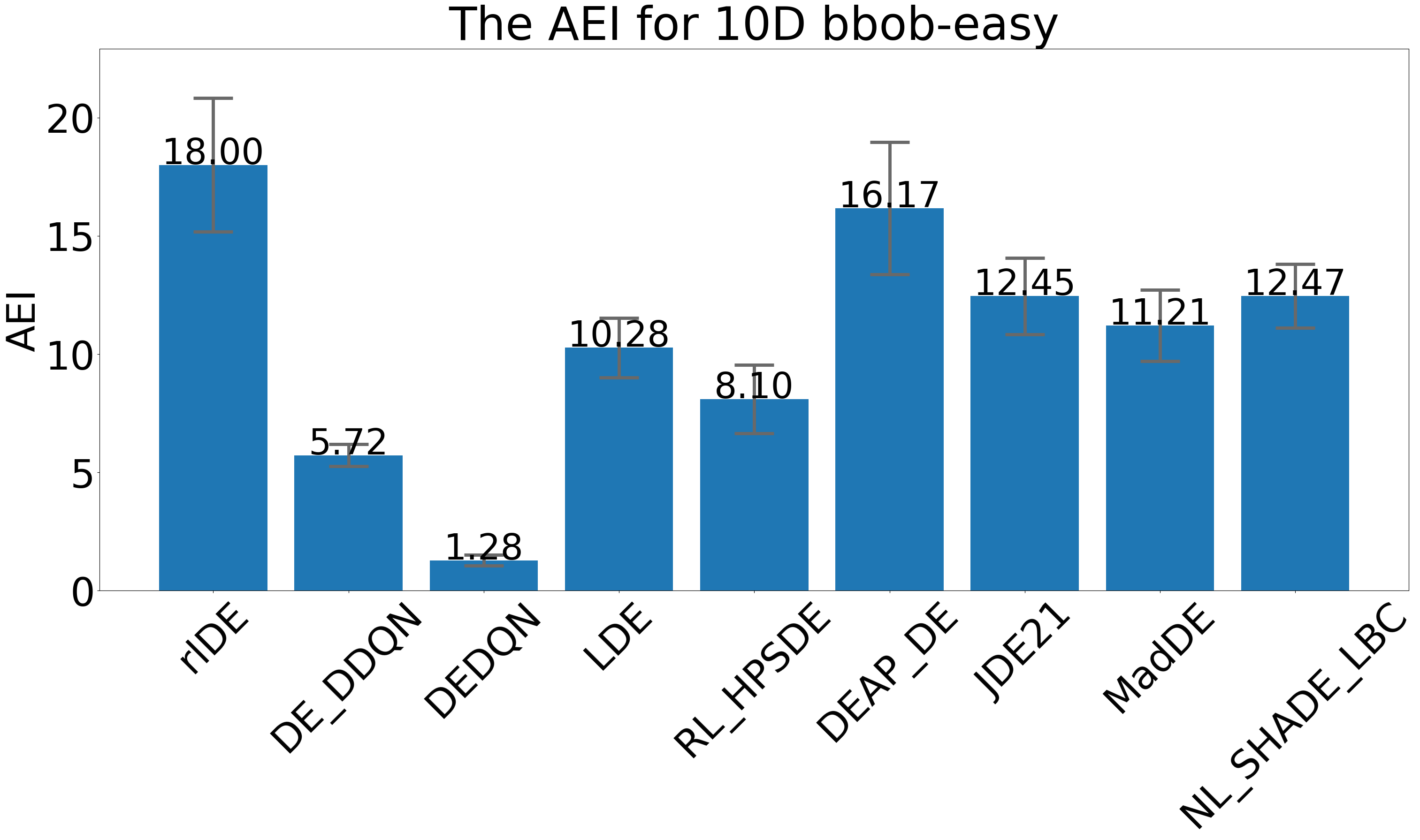}
  \caption{AEI values for BBOB2009 testing functions, where the X-axis represents different algorithms and Y-axis represents AEI values.}
  \label{figAEI}
\end{figure}

AEI results of our test experiments are shown in \fref{figAEI}. The larger, the better. Comprehensively, rlDE is scored 18.00, which ranks first. DEDQN ranks last mainly to due to the worst performance on objectives. The traditional DEs perform better than RL-assisted DEs except rlDE, indicating their better comprehensive performance. 

It can be provisionally concluded that rlDE exhibits superior stability and performance, thereby solidifying its position as a leading algorithm within the tested functions. Future research will involve the development of additional exploration experiments to further validate and enhance its capabilities.

\section{Conclusion}
\label{conclusion}
In this work, we have introduced a novel framework, rlDE, which employs RL for the automatic design of DE algorithms through meta-learning based on problem characteristics. This framework addresses the challenges of black-box optimization by training an RL agent on a diverse set of BBOPs, enabling it to autonomously generate bespoke DE algorithms tailored to specific problems, particularly excelling on unseen problems. The implementation utilizes a DDQN.

The development of rlDE is a response to the limitations of traditional DE design and the need for algorithms capable of adapting to the dynamic and diverse nature of real-world optimization problems. Unlike hyper-heuristic and self-adaptive methods, rlDE directly generates DEs with different components and parameters based on a predefined pool, eliminating the need for additional search processes.

Despite the promising results, rlDE is constrained by the predefined design space, which limits the diversity of the generated DEs. Additionally, the performance of the RL agent is heavily dependent on the breadth and representativeness of the BBOPs it is trained on, which can be both time-consuming and incomplete. Moreover, the RL agent's stability is a concern, and there is a lack of exploration in its application for DE design, such as other architectures and specialized reward functions.

For future research, we plan to investigate a more flexible design space for evolutionary algorithms. We will also develop more efficient and representative methods to explore problem characteristics. Furthermore, we will consider additional strategies to enhance the performance and stability of the RL agent.

\section*{Acknowledgment}
The authors gratefully acknowledge the financial support provided by the Open Project of Xiangjiang Laboratory (No.22XJ02003), the National Science Fund for Outstanding Young Scholars (62122093),the National Natural Science Foundation of China (72421002,62303476), the Science \& Technology Project for Young and Middle-aged Talents of Hunan (2023TJ-Z03), and the University Fundamental Research Fund(23-ZZCX-JDZ-28). The authors would also like to thank the support from COSTA: complex system optimization team of the College of System Engineering at NUDT.


\bibliographystyle{IEEEtran}
\bibliography{COOref}
\begin{IEEEbiography}[{\includegraphics[width=1in,height=1.25in,clip,keepaspectratio]{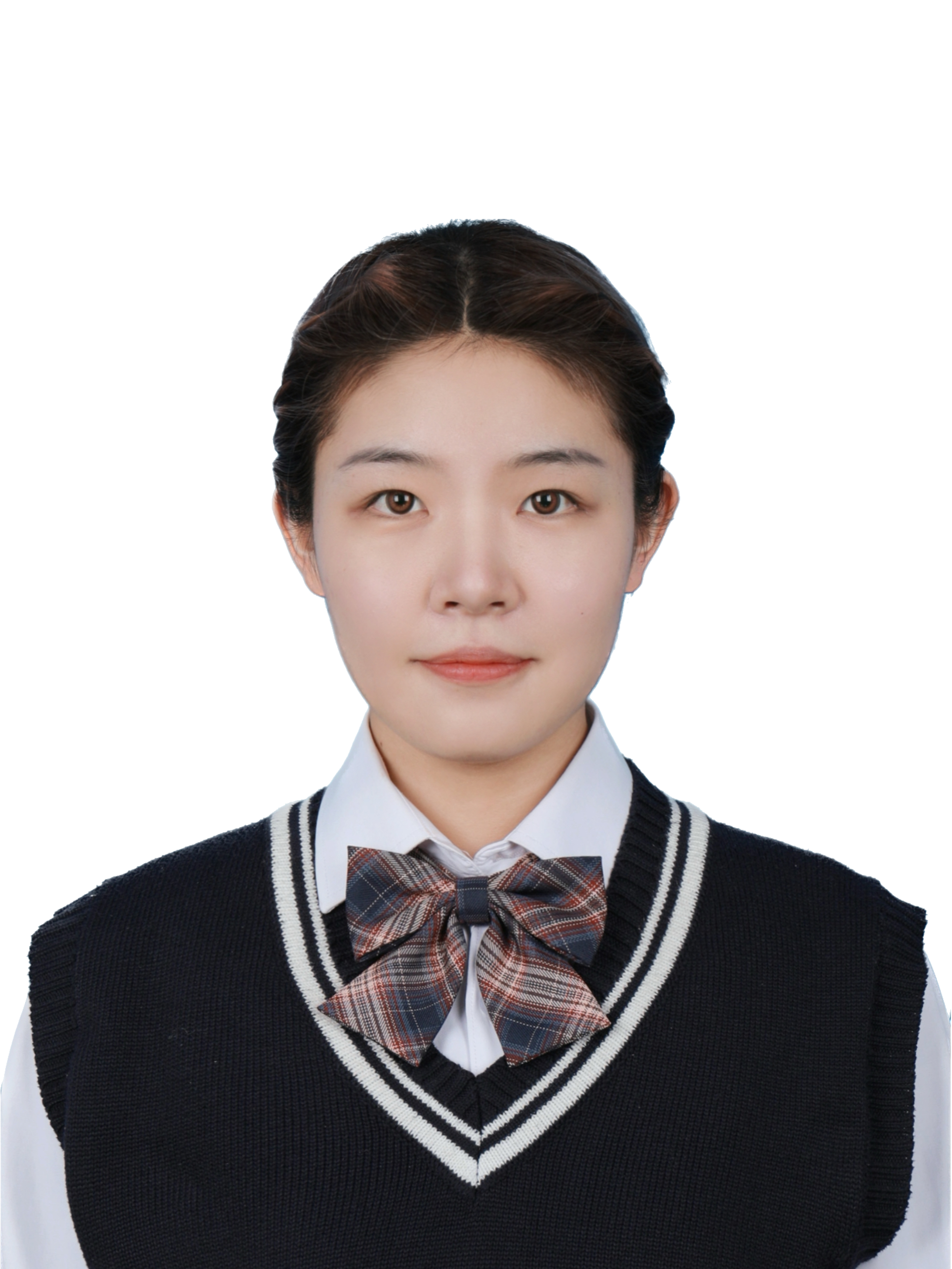}}]{Xu Yang}
    is a PhD student majoring in management science and engineering in National University of Defense Technology (NUDT). She received the M.S. degree from NUDT in 2022. Her research interests include automated algorithm design, computation intelligence, black-box optimization, and optimization methods on Energy Internet.
  \end{IEEEbiography}
  
  \begin{IEEEbiographynophoto}{Rui Wang}
    (Senior Member, IEEE) received the B.S. degree in system engineering from the National University of Defense Technology (NUDT) in 2008, and the Ph.D. degree in system engineering from the University of Sheffield, UK in 2013. Currently, he is with the National University of Defense Technology. His current research interests include evolutionary computation, multiobjective optimization and the development of algorithms applicable in practice. 
    
    Professor Wang has authored more than 40 referred papers including those published in IEEE Transactions on Evolutionary Computation, IEEE Transactions on Cybernetics, and Information Sciences. He serves as an Associate Editor of the IEEE Transactions on Evolutionary Computation, Swarm and Evolutionary Computation, Expert System with Applications, etc. He is the recipients of The Operational Research Society Ph.D. Prize in 2014, of the Funds for Distinguished Young Scientists from the Natural Science Foundation of Hunan province at 2016, of the Wu Wen-Jun Artificial Intelligence Outstanding Young Scholar at 2017, of the National Science Fund for Outstanding Young Scholars at 2021.
  \end{IEEEbiographynophoto}

  \begin{IEEEbiographynophoto}{Kaiwen Li}
    received the B.S., M.S. and Ph.D. degrees in management science and engineering from National University of Defense Technology (NUDT), in 2016, 2018 and 2022, respectively. He is an associate professor with the College of Systems Engineering, NUDT. His research interests include prediction technique, multiobjective optimization, reinforcement learning, data mining, and optimization methods on energy Internet.
  \end{IEEEbiographynophoto}
  
  \begin{IEEEbiographynophoto}{Ling Wang}
    received the B.Sc. degree in automation,and the Ph.D.degree in control theory and control engineering from Tsinghua University in 1995 and1999, respectively. Since 1999, he has been with theDepartment of Automation,Tsinghua Universitywhere he became a Full Professor in 2008. His cur-rent research interests include intelligent optimiza-tion and production scheduling.He was the recipient of the National Natural Science Fund for Distinguished Young Scholars ofChina, the National Natural Science Award (second place) in 2014, the Sci-ence and Technology Award of Beijing City in 2008, and the Natural ScienceAward (first place in 2003, and second place in 2007)nominated by the Min-istry of Education of China.
  \end{IEEEbiographynophoto}

\end{document}